# Relationship between Uncertainty in DNNs and Adversarial Attacks

Mabel Ogonna*, Abigail Adeniran†, Adewale Adeyemo‡

*Rawls College of Business, Texas Tech University, Lubbock, TX 79409, USA

†Department of Computing Science and Mathematics, University of Stirling, Scotland, United Kingdom ‡Department of Electrical and Computer Engineering, Tennessee Technological University, Cookeville, TN 38505, USA

wale.kanmi@gmail.com

*Abstract*—Deep Neural Networks (DNNs) have achieved state-of-the-art results and even outperformed human accuracy in many challenging tasks, leading to DNNs' adoption in a variety of fields including natural language processing, pattern recognition, prediction, and control optimization. However, DNNs are accompanied by uncertainty about their results, causing them to predict an outcome that is either incorrect or outside of a certain level of confidence. These uncertainties stem from model or data constraints, which could be exacerbated by adversarial attacks. Adversarial attacks aim to provide perturbed input to DNNs, causing the DNN to make incorrect predictions or increase model uncertainty. In this review, we explore the relationship between DNN uncertainty and adversarial attacks, emphasizing how adversarial attacks might raise DNN uncertainty.

## I. INTRODUCTION

As the burgeoning demand for intelligence in critical area increases, Deep neural networks (DNNs) have been used in a variety of domains, such as computer vision [1], speech recognition [2], natural language processing [3], machine translation [4] etc., where they have achieved comparable outcomes and in some cases surpassing human expert performance. However, the deployments of DNNs for mission critical applications are limited due to inexplicable DNN inference computation, sensitivity to domain shifts and vulnerability to adversarial attacks [5].

Typically, the input layer of a DNN receive the data and a prediction is made at the output layer after multiple convolution and pooling operations (see Fig. 1). The output of the DNN also displays the confidence score, explaining what a DNN knows and what it doesn't know. Uncertainty in DNNs are broadly divided into data uncertainty (epistemic uncertainty) and model uncertainty (Aleatoric uncertainty) [6]. The epistemic uncertainty explains the uncertainty produced by the model's shortcomings, such as errors in the learning phase, an unsuitable model structure, insufficient information due to unidentified samples, or poor training data set coverage. In contrast, Aleatoric uncertainty is caused by inherent random effects (such as noise) which change the distribution of the data sample from the population [7].

In this review, we discuss the relationship between uncertainty in DNNs and Adversarial attacks. The remainder of the paper is structured as follows: Section II describes Uncertainty in DNNs, Section III describes Adversarial attacks on DNNs, section IV describes Relationship between Uncertainty in DNNs and Adversarial Attacks and Section V concludes this paper.

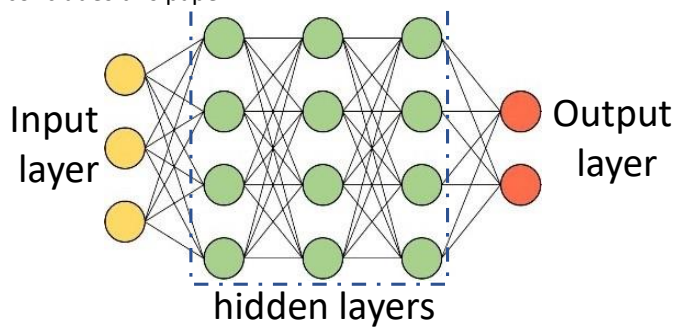

Fig. 1. An architecture of Deep Neural Networks comprising of the input, hidden and output layers.

## II. UNCERTAINTY IN DNNS

The lack of trust in each DNN output is referred to as *uncertainty* [8]. While it is impossible to design a DNN algorithm with 100% confidence, it is imperative to understand what causes uncertainty, how to measure it, and how to minimize it. Since deep learning models don't provide uncertainty estimates and frequently make overly or underly confident predictions, they do in fact need a more thorough evaluation. The research community wants to ensure that DNNs accurately describe the probability that their results will be incorrect or fall outside of a specified range of accuracy.

To make safe and informed judgments, DNNs would need to produce both an output and a degree of certainty in those outcomes. This indicates that the DNNs would convey information about their level of uncertainty and if it is low enough for the output to be trusted along with their findings. In order to handle the decision-making process, the algorithm may require human intervention after producing an output with a high level of uncertainty. Five factors that cause uncertainty in a DNN's predictions include: variability in real-world situations, errors inherent to the measurement systems, errors in the architecture specification of the DNN, errors in the training procedure of the DNN, and errors caused by unknown data.

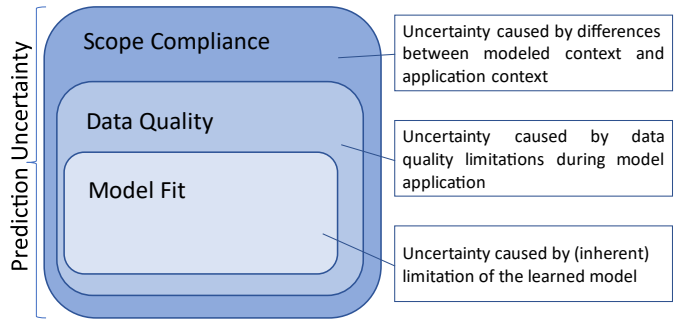

Fig. 2. Onion layer model of uncertainty in DNN application outcomes [9].

## A. Types of Uncertainties

- Aleatoric Uncertainty: This is the uncertainty arising from the natural stochasticity of observations. As aleatoric uncertainty refers to the inherent noise in all observations, it cannot be reduced. Aleatoric uncertainty cannot be reduced even when more data is provided. When it comes to measurement errors, we call it homoscedastic uncertainty because it is constant for all samples. Input data-dependent uncertainty is known as heteroscedastic uncertainty.
- Epistemic Uncertainty: This describes what the model does not know because of the lack of training data. Epistemic uncertainty is due to limited data and knowledge. Epistemic uncertainty is reducible, which means it can be lowered by providing additional data. Epistemic uncertainty can arise in areas where there are fewer samples for training.

## B. Prediction uncertainty

"Prediction uncertainty" is the uncertainty that is conveyed in the model's output and is determined by summing the epistemic uncertainty and the aleatoric uncertainty [10].

### III. ADVERSARIAL ATTACKS ON DNNs

Adversarial attacks on DNNs involves the addition of subtle malicious perturbation on clean images to fool the DNN model and cause misclassification as shown in Fig. 3. The vulnerability of DNNs to adversarial attacks was first noted by Szegedy et al., [11] and Goodfellow et al., [12], after which several other researchers have proposed new adversarial attack methods and defense methods. Adversarial attacks could be targeted or non-targeted. Targeted attacks attempt to force the model to predict an input as a specified target class which differs from the true class, while non-targeted attacks only attempt to enforce the model to misclassify the input. In literature, the three types of adversarial attacks include: white box, black box, and gray box attacks.

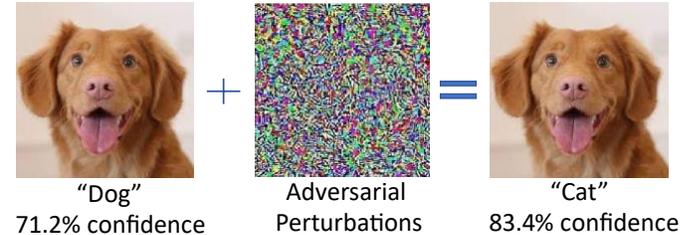

Fig. 3. A DNN model considers the original image as a "dog" (71.2%). However, when a small amount of perturbation (that is imperceptible to the human eyes) is added, the same model classifies the image as a "cat"(83.4%) [12]

- White box attack: this type of attack assumes that the attacker is privy to the model's parameters which include the architecture, weights, and the dataset. The knowledge of the model parameters has been found to be useful in designing imperceptible adversarial perturbations which can reduce the DNNs performance [11]. Examples include the Fast gradient sign Method (FGSM), Iterative fast gradient sign method (I-FGSM), etc.
- Black box attack: this type of attack assumes that the attacker has no knowledge of model parameters. Adversarial examples are either generated via adversarial transferability or model inversion technique.
- Gray box attack: this type of attack assumes the attacker has partial knowledge of the model parameters.

### IV. RELATIONSHIP BETWEEN UNCERTAINTY IN DNNs AND ADVERSARIAL ATTACKS

Adversarial machine learning attacks are premised on perturbing the input to a DNN in a way that increases the likelihood of incorrect decision-making and results in false predictions [13]. Adversarial attacks increase uncertainty in DNNs (especially Epistemic uncertainty). Adversarial attacks change the pixel distribution of an input image, causing the DNN to suffer from either over-confidence or underconfidence predictions. The uncertainty of a DNN model can be used to infer the goal of an adversary. Adversarial goals can be classified into four groups based on how they affect the DNN's output integrity, i.e., confidence reduction, misclassification, targeted misclassification, and source/target misclassification [14].

- Confidence Reduction: The adversary's goal is to reduce the target model's prediction confidence, which increases

the DNN's uncertainty. The adversarial sample of a "dog," for example, is predicted with lesser confidence
- Misclassification: The adversaries attempt to change the output classification of input to any class other than the original class. Even if the classification is incorrect, the prediction score for the misclassified input may be high. For example, an adversarial sample of a "dog" is highly confidently predicted to be a cat as seen in Fig. 3.
- Targeted Misclassification: The adversaries attempt to modify the output classification of the *input* to a *specific target class*, for example, regardless of any adversarial samples inputted into a classifier, it is predicted to be a "cat".
- Source/Target Misclassification: The adversaries attempt to change the output classification of a *special input* to a *special target class*, e.g., only "dog" class is predicted to be a "cat".

## V. Conclusion

There is some degree of uncertainty in every machinelearning project. While some uncertainty will always exist, there are several strategies for identifying, calculating, and minimizing it. Adversarial attacks, on the other hand, increase the uncertainty associated with a DNN model by changing the distribution of the pixels in the input image (epistemic uncertainty), thus confusing the DNN model. it is imperative for developers to create redundancy solutions that ensure DNNs deployed to production meet Service Level Agreements, particularly for mission-critical systems.

## References

[1] L. G. Shapiro, G. C. Stockman *et al.*, *Computer vision*. Prentice Hall New Jersey, 2001, vol. 3.

[2] O. Abdel-Hamid, A.-r. Mohamed, H. Jiang, L. Deng, G. Penn, and D. Yu, "Convolutional neural networks for speech recognition," *IEEE/ACM Transactions on audio, speech, and language processing*, vol. 22, no. 10, pp. 1533–1545, 2014.

[3] K. Chowdhary, "Natural language processing," *Fundamentals of artificial intelligence*, pp. 603–649, 2020.

[4] P. T. Krishnan and P. Balasubramanian, "Detection of alphabets for machine translation of sign language using deep neural net," in *2019 International Conference on Data Science and Communication (IconDSC)*. IEEE, 2019, pp. 1–3.

[5] J. Gawlikowski, C. R. N. Tassi, M. Ali, J. Lee, M. Humt, J. Feng, A. Kruspe, R. Triebel, P. Jung, R. Roscher *et al.*, "A survey of uncertainty in deep neural networks," *arXiv preprint arXiv:2107.03342*, 2021.

[6] Y. Gal and Z. Ghahramani, "Dropout as a bayesian approximation: Representing model uncertainty in deep learning," in *international conference on machine learning*. PMLR, 2016, pp. 1050–1059.

[7] D. Feng, L. Rosenbaum, and K. Dietmayer, "Towards safe autonomous driving: Capture uncertainty in the deep neural network for lidar 3d vehicle detection," in *2018 21st international conference on intelligent transportation systems (ITSC)*. IEEE, 2018, pp. 3266–3273.

[8] X. Zhang, X. Xie, L. Ma, X. Du, Q. Hu, Y. Liu, J. Zhao, and M. Sun, "Towards characterizing adversarial defects of deep learning software from the lens of uncertainty," in *2020 IEEE/ACM 42nd International Conference on Software Engineering (ICSE)*. IEEE, 2020, pp. 739–751.

[9] M. Klas and A. M. Vollmer, "Uncertainty in machine learning applications: A practice-driven classification of uncertainty," in *International conference on computer safety, reliability, and security*. Springer, 2018, pp. 431–438.

[10] H. Asgharnezhad, A. Shamsi, R. Alizadehsani, A. Khosravi, S. Nahavandi, Z. A. Sani, D. Srinivasan, and S. M. S. Islam, "Objective evaluation of deep uncertainty predictions for covid-19 detection," *Scientific Reports*, vol. 12, no. 1, pp. 1–11, 2022.

[11] C. Szegedy, W. Zaremba, I. Sutskever, J. Bruna, D. Erhan, I. Goodfellow, and R. Fergus, "Intriguing properties of neural networks," *arXiv preprint arXiv:1312.6199*, 2013.

[12] A. Kurakin, I. J. Goodfellow, and S. Bengio, "Adversarial examples in the physical world," in *Artificial intelligence safety and security*. Chapman and Hall/CRC, 2018, pp. 99–112.

[13] O. F. Tuna, F. O. Catak, and M. T. Eskil, "Exploiting epistemic uncertainty of the deep learning models to generate adversarial samples," *Multimedia Tools and Applications*, vol. 81, no. 8, pp. 11479–11500, 2022.

[14] S. Qiu, Q. Liu, S. Zhou, and C. Wu, "Review of artificial intelligence adversarial attack and defense technologies," *Applied Sciences*, vol. 9, no. 5, p. 909, 2019.